\documentclass[11pt,a4paper]{article}
\usepackage[hyperref]{acl2019}
\usepackage{times}
\usepackage{latexsym}

\usepackage{url}

\usepackage{amsfonts, bm} 
\usepackage{amsmath} 
\usepackage{float}
\usepackage{array}
\usepackage{booktabs} 
\usepackage[english]{babel}
\usepackage{multirow,graphicx,paralist}
\usepackage{makecell}

\def\aclpaperid{1st NLP4ConvAI Workshop} 

\title{FlowDelta: Modeling Flow Information Gain in Reasoning\\ for Conversational Machine Comprehension}

\aclfinalcopy
\author{Yi-Ting Yeh \\
National Taiwan University \\
\texttt{r07922064@csie.ntu.edu.tw} \\\And
Yun-Nung Chen \\
National Taiwan University \\
\texttt{y.v.chen@ieee.org} \\}

\date{}

\begin{document}
\maketitle
\begin{abstract}
  Conversational machine comprehension requires deep understanding of the dialogue flow, and the prior work proposed FlowQA to implicitly model the context representations in reasoning for better understanding.
  This paper proposes to explicitly model the \emph{information gain} through dialogue reasoning in order to allow the model to focus on more informative cues.
  The proposed model achieves state-of-the-art performance in a conversational QA dataset QuAC and sequential instruction understanding dataset SCONE, which shows the effectiveness of the proposed mechanism and demonstrates its capability of generalization to different QA models and tasks \footnote{Our code can be found in \url{https://github.com/MiuLab/FlowDelta}.}

\end{abstract}

\section{Introduction}

Machine reading comprehension has been increasingly studied in the NLP area, which aims to read a given passage and then answer questions correctly. 
However, human usually seeks answers in a conversational manner by asking follow-up questions given the previous answers.
Traditional machine reading comprehension (MC) tasks such as SQuAD \cite{rajpurkar2016squad} focus on a single-turn setting, and there is no connection between different questions and answers to the same passage. 
To address the multi-turn issue, several datasets about conversational question answering (QA) were introduced, such as CoQA~\cite{reddy2018coqa} and QuAC~\cite{choi2018quac}.

\begin{figure}[t!]
  \centering
  \includegraphics[width=.95\linewidth]{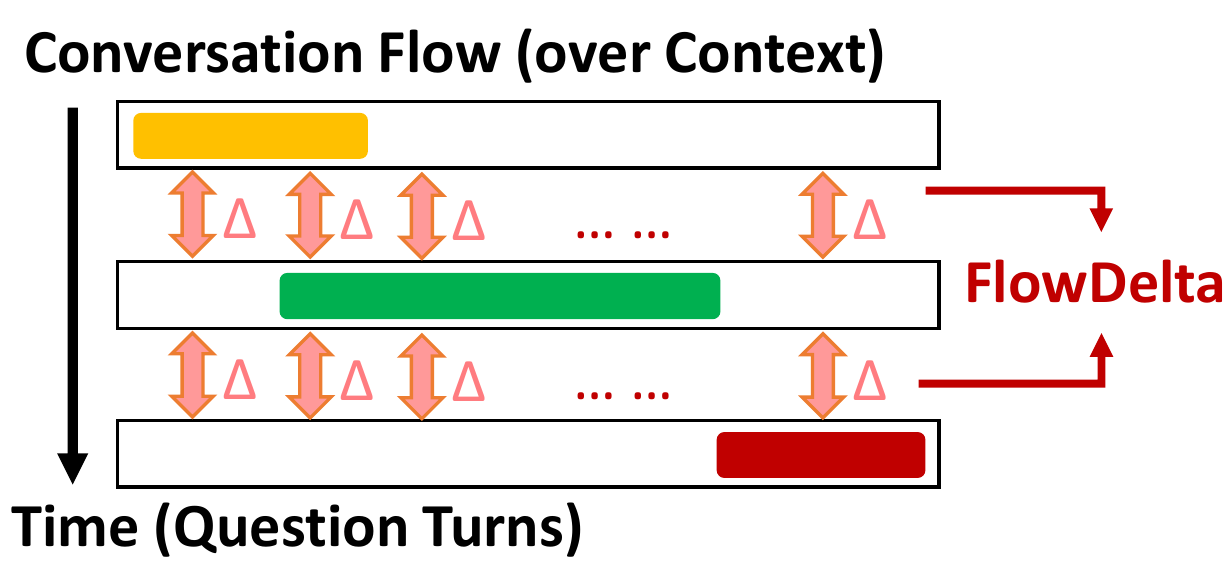}
  \vspace{-1mm}
  \caption{Illustration of the flow information gain modeled by the FlowDelta mechanism.}
  \label{fig:idea}
  \vspace{-5mm}
\end{figure}

Most existing machine comprehension models apply single-turn methods and augment the input with question and answer history, ignoring previous reasoning processes in the models.
Recently proposed FlowQA~\cite{huang2018flowqa} attempted at modeling such multi-turn reasoning in dialogues in order to improve performance for conversational QA.
However, the proposed \textsc{Flow} operation is expected to incorporate salient information in an \emph{implicit} manner, because the learned representations captured by \textsc{Flow} would change during multi-turn questions.
It is unsure whether such change correlates well with the current answer or not.
In order to \emph{explicitly} model the information gain in \textsc{Flow} and further relate the current answer to the corresponding context, we present a novel mechanism, FlowDelta, which focuses on modeling the difference between the learned context representations in multi-turn dialogues illustrated in Figure~\ref{fig:idea}.
The contributions are 3-fold:
\begin{compactitem}
\item This paper proposes a simple and effective mechanism to explicitly model information gain in flow-based reasoning for multi-turn dialogues, which can be easily incorporated in different MC models.
\item FlowDelta consistently improves the performance on various conversational MC datasets, including CoQA and QuAC.
\item The proposed method achieves the state-of-the-art results among published models on QuAC and sequential instruction understanding task (SCONE).
\end{compactitem}

\begin{figure*}[t!]
  \centering
  \includegraphics[width=.9\linewidth]{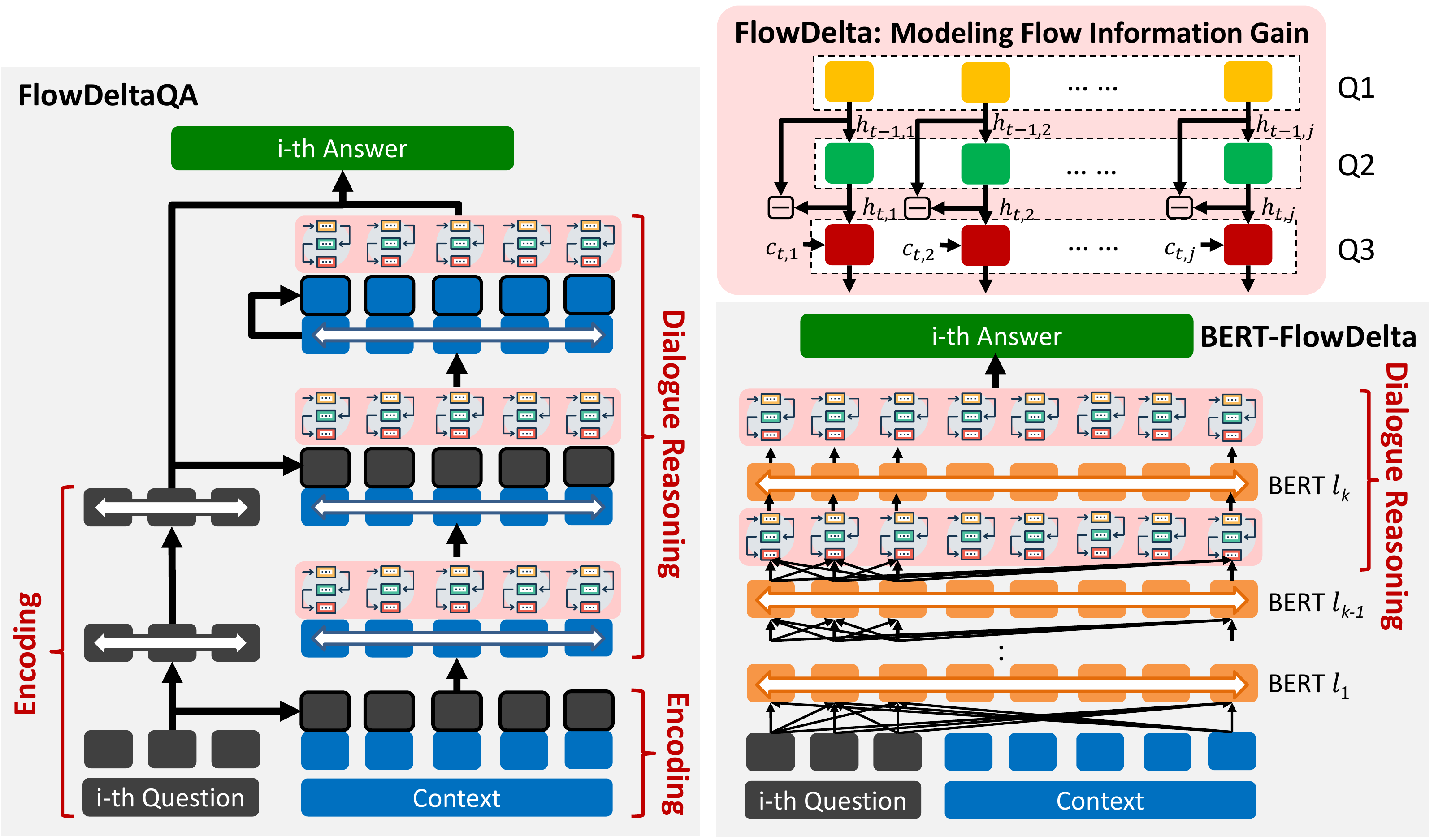}
  \vspace{-2mm}
  \caption{Illustration of the proposed FlowDelta models.}
  \label{fig:model}
  \vspace{-4mm}
\end{figure*}

\section{Background} \label{sec:background}


Given a document (context), previous conversation history (i.e., question/answer pairs) and the current question, the goal of conversational QA is to find the correct answer. 
We denote the context document as a sequence of $m$ words $\boldsymbol{C} = \{c_1, c_2, \ldots. c_m\}$, and the $i$-th question $\boldsymbol{Q}_i = \{q_1, q_2, \ldots, q_n\}$ as a sequence of $n$ words. 
In the extractive setting, the $i$-th answer $\boldsymbol{A}_i$ is guaranteed to be a span in the context. 
The main challenge in conversational QA is that current question may depend on the conversation history, which differs from the classic machine comprehension. 
Therefore, how to incorporate previous history into the QA model is especially important for better understanding.
Prior work~\cite{huang2018flowqa} proposes an effective way to model the reasoning in multi-turn dialogues summarized below.

\paragraph{\textsc{Flow} Operation}
Instead of only using shallow history like previous questions and answers, \citet{huang2018flowqa} proposed the \textsc{Flow} operation that feeds the model with entire hidden representations generated during the reasoning process when answering previous questions.
\textsc{Flow} is defined as \emph{a sequence of latent representations based on the context tokens} and is demonstrated effective for conversational QA tasks, because it well incorporates  multi-turn information in dialogue reasoning.

Let the context representation for $i$-th question be $\boldsymbol{C}_i = c_{i,1}, \ldots, c_{i,m}$ and the dialogue length is $t$. 
When answering questions in the dialogue, there are $t$ context sequences of length $m$, one for each question. 
We reshape it to become $m$ sequences of length $t$, one for each context word, and then pass each sequence into a unidirectional GRU. All context word representation $j$  ($1 \leq j \leq m$) are processed in parallel in order to model the information via the \textsc{Flow} direction (vertical direction illustrated in Figure~\ref{fig:idea}).
\begin{equation}
    \label{eq:flow}
     h_{1, j}, \ldots, h_{t, j}  = GRU(c_{1, j}, \ldots, c_{t, j}) 
\end{equation}
Then we reshape the outputs from GRU back and form $F_i = \{h_{i, 1}, \ldots, h_{i, m}\}$, where $F_i$ is the output of the \textsc{Flow} layer.

\paragraph{FlowQA}
The \textsc{Flow} layer described above is incorporated in \textsc{FlowQA} for conversational MC, which is built on the single-turn MC model FusionNet~\cite{huang2017fusionnet}, and the full structure is shown in the left part of Figuire~\ref{fig:model}. 
Briefly, \textsc{FlowQA} first performs word-level attention to fuse the information of $i$-th question $\boldsymbol{Q}_i$ into context $\boldsymbol{C}$. 
Then it uses two LSTM cells combined with \textsc{Flow} layers to integrate the context representations, followed by the context-question attention computation.
Finally, \textsc{FlowQA} performs self-attention \cite{yu2018qanet} on the context and predict the answer span.
Modeling \textsc{Flow} is shown effective to improve the performance for conversational MC.

\section{Proposed Approaches}

This paper extends the concept of \textsc{Flow} and proposes a flow-based approach,  \textsc{FlowDelta}, to \emph{explicitly} model information gain in flow during dialogues illustrated in Figure~\ref{fig:model}.
The proposed mechanism is flexible to integrate with different models, including FlowQA and others.
To examine such flexibility and generalization capability, we further apply \textsc{Flow} and \textsc{FlowDelta} to BERT \cite{devlin2018bert}, a pretrained language understanding model that shows strong performance in MC tasks, to allow model to grasp dialogue history.

\subsection{FlowDeltaQA}

In the original \textsc{Flow} operation in (\ref{eq:flow}), the $k$-th step computation of GRU is $h_{k, j} = GRU(c_{k, j}, h_{k-1, j})$. 
We assume that the difference of previous hidden representations $h_{k-1, j}$ and $h_{k-2, j}$ indicates whether the flow change is important, which can be viewed as the information gain through the reasoning process.
For example, 3 consecutive questions $Q_{k-2}, Q_{k-1}, Q_{k}$. $Q_{k-1}$ and $Q_{k}$ all discuss the same event described in the span $\{c_j, c_{j+1}, \ldots, c_l\}$ of the context, while $Q_{k-2}$ is about another topic. 
We expect the hidden state $\{h_{k-1, j}, h_{k-1, j+1} \ldots, h_{k-1, l}\}$ of the span in turn $k-1$ is dissimilar to the hidden state in the turn $k-2$, because their topics are different.
By explicitly modeling such difference, our model more easily relates the current reasoning process to the corresponding context. 

Following the intuition above, we propose \textsc{FlowDelta} by modifying the single step computation of \textsc{Flow} into: 
\begin{eqnarray}
    \label{eq: flowdelta}
    h_{k, j} = GRU([c_{k, j}; h_{k-1, j} - h_{k-2, j}], h_{k-1, j}),
\end{eqnarray}
where $[x;y]$ is the concatenation of the vectors $x$ and $y$.
We also investigate other variants such as Hadamard product ($h_{k-1, j} * h_{k-2, j}$) detailed in Appendix \ref{sec: flow_variant}.

\subsection{BERT-FlowDelta}

BERT~\cite{devlin2018bert} with fine-tuning recently has reached the state-of-the-art in many single-turn MC tasks, such as SQuAD~\cite{rajpurkar2016squad,rajpurkar2018know}. 
However, how to extend BERT to the multi-turn setting remains unsolved.
We propose to incorporate the \textsc{FlowDelta} mechanism to deal with the multi-turn problem, where the \textsc{Flow} layer automatically integrates multi-turn information instead of tuning the number of QA pairs for inclusion.

Each layer of BERT is a Transformer block \cite{vaswani2017attention} that consists of multi-head attention (MH) and fully-connected feed forward network (FFN):
\begin{eqnarray}
    \label{eq:bert}
    h_{l+1} = \text{Transfomer}(h_l)= LN(h_l + SA(h_l)),\nonumber\\
    SA(h) = FFN(LN(h + MH(h)),\nonumber
\end{eqnarray}
where $h_l$ is the hidden representation of the $l$-th layer, LN is layer normalization \cite{ba2016layer} and SA means self-attention.
To utilize $L$ layers from BERT for the extractive question answering task, we feed the hidden representation from last layer $h_L$ to a fully-connected layer (NN) to predict the answer span, written as $P^S, P^E = NN(h_L)$, where $P^S$ and $P^E$ are span start and span end probability for each word respectively.

BERT-FlowDelta incorporates the proposed \textsc{FlowDelta} mechanisms for two parts shown in the bottom right corner of Figure \ref{fig:model}.
First, we add \textsc{FlowDelta} layer before the final prediction layer, $P^S, P^E = NN([h_L; \text{FlowDelta}(h_L)])$. 
Second, we further insert \textsc{FlowDelta} into the last BERT layer, considering that modeling dialogue history \emph{within} BERT may be benefitial.
\begin{eqnarray}
    \label{eq:bert_flowdelta}
    h_L = LN(h_{L-1} + SA(h_{L-1}) + \text{FlowDelta}(h_{L-1}))\nonumber
\end{eqnarray}
These two modifications are called exFlowDelta and inFlowDelta respectively, and the latter
 also meets the idea from \citeauthor{stickland2019bert} who added additional parameters into BERT layers to improve the performance of multi-task learning.
In our experiments, we only modify the last BERT layer to avoid largely increasing model size.

\section{Experiments}\label{sec:exp}
To evaluate the effectiveness of the proposed \textsc{FlowDelta}, various tasks that contains dialogue history for understanding are performed in the following experiments.

\begin{table*}[t!]
    \begin{center}
    \small
    \begin{tabular}{|l|c|cc|c|ccc|}
    \hline
    \multirow{2}{*}{\bf Model} & \multicolumn{3}{c|}{\bf CoQA}  & \multicolumn{4}{c|}{\bf QuAC}\\
    & \bf Dev & \multicolumn{2}{c|}{\bf Test} & \bf Dev & \multicolumn{3}{|c|}{\bf Test}\\
    \cline{2-8}
       & {\bf F1} &{\bf Child/Liter/Mid/News/Wiki/Reddit/Sci} &  {\bf F1} & {\bf F1} & {\bf F1} & {\bf HEQ-Q} & {\bf HEQ-D}\\
        \hline\hline
        BiDAF++ (N-ctx) & 69.2 &  66.5~~65.7~~70.2~~71.6~~72.6~~60.8~~67.1 & 67.8 & 60.6 &60.1 & 54.8 & 4.0 \\
        FlowQA & 76.7 & 73.7~~71.6~~76.8~~79.0~~80.2~~67.8~~76.1 & 75.0 & 63.9 & 64.1 & 59.6 & 5.8\\
        SDNet \cite{zhu2018sdnet} & 78.0 & 75.4~~73.9~~77.1~~80.3~~\textbf{83.1}~~69.8~~76.8 & 76.6 & - & - & - & -\\ 
        HAM \cite{qu2019attentive} & - & - & - & - & 65.4 & \bf 61.8 & 6.7 \\
        \hline
        FlowDeltaQA  & 77.6 & - & - & 64.8 & - & - & - \\
        BERT-FlowDelta &\textbf{79.4} & \textbf{75.9}~~\textbf{ 75.6}~~\textbf{80.1}~~\textbf{82.1}~~82.3~~\textbf{69.8}~~\textbf{78.8} & \bf 77.7  & \bf 66.1 & \bf 65.5 &  61.0 & \bf 6.9 \\
        \hline
        Human  & 89.8 & 90.2~~88.4~~89.8~~88.6~~89.9~~86.7~~88.1&88.8 & 80.8 & 81.1 & 100 & 100\\
    \hline
    \end{tabular}
    \end{center}
    \vspace{-3mm}
    \caption{\label{tab:main} Conversational QA results on CoQA and QuAC, where (N-ctx) refers to using previous N QA pairs (\%).}
    \vspace{-2mm}
\end{table*}

\begin{table}[t!]
    \begin{center}
    \small
    \begin{tabular}{|l|c|c|}
    \hline
\bf Model &\bf CoQA F1 & \bf QuAC F1\\
    \hline\hline
        BERT-FlowDelta & \bf 79.4 & \bf 66.1 \\
        - inFlowDelta & 79.0 & 64.1 \\
        ~~ - exFlowDelta & 78.0 & 62.3 \\
    \hline
    BERT-Flow & 79.2 & 64.3 \\
    \hline
    \end{tabular}
    \end{center}
    \vspace{-3mm}
    \caption{\label{tab: bert_flow_ablation} The ablation study of BERT-FlowDelta (\%).}
    \vspace{-2mm}
\end{table}

\begin{table}[t!]
    \begin{center}
    \small
    \begin{tabular}{|l|ccc|}
    \hline
        {\bf Model} & {\bf Scene} & {\bf Tangrams} & {\bf Alchemy} \\
    \hline\hline
        \citet{long2016simpler} & 14.7 & 27.6 & 52.3 \\
        \citet{guu2017language} & 46.2 & 37.1 & 52.9 \\
        \citet{suhr2018situated} & 66.4 & 60.1 & 62.3 \\
        \citet{fried2017unified} & 72.7 & 69.6 & 72.0 \\
        FusionNet & 58.2 & 67.9 & 74.1 \\
        FlowQA & 74.5 & 72.3 & \bf 76.4 \\
        FlowDeltaQA & \bf 75.1 & \bf 72.5 & 76.1\\
    \hline
    \end{tabular}
    \end{center}
    \vspace{-3mm}
    \caption{\label{tab: scone_main} Dialogue accuracy for SCONE test (\%). }
    \vspace{-3mm}
\end{table}

\subsection{Setup}

Our models are tested on two conversational MC datasets, CoQA~\cite{reddy2018coqa} and QuAC~\cite{choi2018quac}, and a sequential instruction understanding dataset, SCONE~\cite{long2016simpler}.
For QuAC, we also report the Human Equivalence Score (HEQ). 
HEQ-Q and HEQ-D represent the percentage of exceeding the model performance over the human evaluation for each question and dialogue respectively.
While CoQA and QuAC both follow the conversational QA setting, SCONE is the task requiring model to understand a sequence of natural language instructions and modify the word state accordingly. 
We follow \citet{huang2018flowqa} to reduce instruction understanding to machine comprehension. 
Appendix \ref{sec:scone_reduce} contains the example and reduction detail of SCONE for reference.

\subsection{Results}

Table \ref{tab:main} reports model performance on CoQA and QuAC.
It can be found that FlowDeltaQA yields substantial improvement over FlowQA on both datasets (+ 0.9 $\%$ F1 on both CoQA and QuAC), showing the usefulness of explicitly modeling the information gain in the \textsc{Flow} layer.
Furthermore, BERT-FlowDelta outperforms the published models on QuAC leaderboard on Apr 24, 2019.
Specifically, while BERT-FlowDelta achieves slightly worse HEQ-Q score on QuAC to the HAM model \cite{qu2019attentive}, we outperform HAM in HEQ-D metrics, showing the superiority of our model in modeling whole dialogue.
Note that \textsc{FlowDelta} actually introduced few additional parameters compared to \textsc{Flow}, since it only augments the input dimension of GRU.
The consistent improvement from both data demonstrates the generalization capability of applying the proposed mechanism to various models.

Table \ref{tab: bert_flow_ablation} shows the ablation study of BERT-FlowDelta, where two proposed modules are both important for achieving such results.
It is interesting that the proposed inFlowDelta and exFlowDelta boost the performance more on QuAC.
As \citet{yatskar2018qualitative} mentioned, the topics in a dialogue shift more frequently on QuAC than on CoQA, and we can see vanilla BERT also performs well on CoQA in the ablation of \textsc{Flow} which provides long term dialog history information.
Therefore, we can conclude that while \textsc{FlowDelta} improves the ability to grasp information gain in the dialog, it bring less performance improvement in the setting we do not need much contexts to answer the question.

Table \ref{tab: scone_main} shows the performance of our FlowDeltaQA on the SCONE
\footnote{The results of BERT-FlowDelta are not shown, since SCONE is a relatively small and synthetic dataset.}.
Our model outperforms FlowQA and achieves the state-of-the-art in \textsc{Scene} and \textsc{Tangrams} domains. 
The small performance drop in \textsc{Alchemy} aligns well with the statement in the ablation study.
Because experiments show that removing \textsc{Flow} affects performance in \textsc{Alchemy} less when comparing between FlowQA and FusionNet~\cite{huang2017fusionnet} (same models except \textsc{Flow}), we claim that the previous dialogue history is less important in this domain.
Thus replaying \textsc{Flow} with FlowDelta does not bring any improvement in the \textsc{Alchemy} domain. 
The detailed qualitative study can be found in Appendix~\ref{sec:qual}.

\section{Conclusion}
This paper presents a simple and effective extension of \textsc{Flow} named \textsc{FlowDelta}, which is capable of explicitly modeling the dialogue history in reasoning for better conversational machine comprehension. 
The proposed FlowDelta is flexible to apply to other machine comprehension models including FlowQA and BERT.
The experiments on three datasets show that the proposed mechanism can model the information flow in the multi-turn dialogues more comprehensively, and further boosts the performance consistently. 
In the future, we will investigate more efficient ways to model the dialogue flow for conversational tasks.

\bibliography{acl2019}

\begin{thebibliography}{18}
\expandafter\ifx\csname natexlab\endcsname\relax\def\natexlab#1{#1}\fi

\bibitem[{Ba et~al.(2016)Ba, Kiros, and Hinton}]{ba2016layer}
Jimmy~Lei Ba, Jamie~Ryan Kiros, and Geoffrey~E Hinton. 2016.
\newblock Layer normalization.
\newblock \emph{arXiv preprint arXiv:1607.06450}.

\bibitem[{Choi et~al.(2018)Choi, He, Iyyer, Yatskar, Yih, Choi, Liang, and
  Zettlemoyer}]{choi2018quac}
Eunsol Choi, He~He, Mohit Iyyer, Mark Yatskar, Wen-tau Yih, Yejin Choi, Percy
  Liang, and Luke Zettlemoyer. 2018.
\newblock Quac: Question answering in context.
\newblock In \emph{Proceedings of the 2018 Conference on Empirical Methods in
  Natural Language Processing}, pages 2174--2184.

\bibitem[{Devlin et~al.(2018)Devlin, Chang, Lee, and
  Toutanova}]{devlin2018bert}
Jacob Devlin, Ming-Wei Chang, Kenton Lee, and Kristina Toutanova. 2018.
\newblock Bert: Pre-training of deep bidirectional transformers for language
  understanding.
\newblock \emph{arXiv preprint arXiv:1810.04805}.

\bibitem[{Fried et~al.(2017)Fried, Andreas, and Klein}]{fried2017unified}
Daniel Fried, Jacob Andreas, and Dan Klein. 2017.
\newblock Unified pragmatic models for generating and following instructions.
\newblock \emph{arXiv preprint arXiv:1711.04987}.

\bibitem[{Guu et~al.(2017)Guu, Pasupat, Liu, and Liang}]{guu2017language}
Kelvin Guu, Panupong Pasupat, Evan~Zheran Liu, and Percy Liang. 2017.
\newblock From language to programs: Bridging reinforcement learning and
  maximum marginal likelihood.
\newblock \emph{arXiv preprint arXiv:1704.07926}.

\bibitem[{Huang et~al.(2018)Huang, Choi, and Yih}]{huang2018flowqa}
Hsin-Yuan Huang, Eunsol Choi, and Wen-tau Yih. 2018.
\newblock Flowqa: Grasping flow in history for conversational machine
  comprehension.
\newblock \emph{arXiv preprint arXiv:1810.06683}.

\bibitem[{Huang et~al.(2017)Huang, Zhu, Shen, and Chen}]{huang2017fusionnet}
Hsin-Yuan Huang, Chenguang Zhu, Yelong Shen, and Weizhu Chen. 2017.
\newblock Fusionnet: Fusing via fully-aware attention with application to
  machine comprehension.
\newblock \emph{arXiv preprint arXiv:1711.07341}.

\bibitem[{Long et~al.(2016)Long, Pasupat, and Liang}]{long2016simpler}
Reginald Long, Panupong Pasupat, and Percy Liang. 2016.
\newblock Simpler context-dependent logical forms via model projections.
\newblock \emph{arXiv preprint arXiv:1606.05378}.

\bibitem[{Qu et~al.(2019)Qu, Yang, Qiu, Zhang, Chen, Croft, and
  Iyyer}]{qu2019attentive}
Chen Qu, Liu Yang, Minghui Qiu, Yongfeng Zhang, Cen Chen, W~Bruce Croft, and
  Mohit Iyyer. 2019.
\newblock Attentive history selection for conversational question answering.
\newblock In \emph{Proceedings of the 28th ACM International Conference on
  Information and Knowledge Management}, pages 1391--1400.

\bibitem[{Rajpurkar et~al.(2018)Rajpurkar, Jia, and Liang}]{rajpurkar2018know}
Pranav Rajpurkar, Robin Jia, and Percy Liang. 2018.
\newblock Know what you don't know: Unanswerable questions for squad.
\newblock \emph{arXiv preprint arXiv:1806.03822}.

\bibitem[{Rajpurkar et~al.(2016)Rajpurkar, Zhang, Lopyrev, and
  Liang}]{rajpurkar2016squad}
Pranav Rajpurkar, Jian Zhang, Konstantin Lopyrev, and Percy Liang. 2016.
\newblock Squad: 100,000+ questions for machine comprehension of text.
\newblock \emph{arXiv preprint arXiv:1606.05250}.

\bibitem[{Reddy et~al.(2018)Reddy, Chen, and Manning}]{reddy2018coqa}
Siva Reddy, Danqi Chen, and Christopher~D Manning. 2018.
\newblock Coqa: A conversational question answering challenge.
\newblock \emph{arXiv preprint arXiv:1808.07042}.

\bibitem[{Stickland and Murray(2019)}]{stickland2019bert}
Asa~Cooper Stickland and Iain Murray. 2019.
\newblock Bert and pals: Projected attention layers for efficient adaptation in
  multi-task learning.
\newblock \emph{arXiv preprint arXiv:1902.02671}.

\bibitem[{Suhr and Artzi(2018)}]{suhr2018situated}
Alane Suhr and Yoav Artzi. 2018.
\newblock Situated mapping of sequential instructions to actions with
  single-step reward observation.
\newblock \emph{arXiv preprint arXiv:1805.10209}.

\bibitem[{Vaswani et~al.(2017)Vaswani, Shazeer, Parmar, Uszkoreit, Jones,
  Gomez, Kaiser, and Polosukhin}]{vaswani2017attention}
Ashish Vaswani, Noam Shazeer, Niki Parmar, Jakob Uszkoreit, Llion Jones,
  Aidan~N Gomez, {\L}ukasz Kaiser, and Illia Polosukhin. 2017.
\newblock Attention is all you need.
\newblock In \emph{Advances in Neural Information Processing Systems}, pages
  5998--6008.

\bibitem[{Yatskar(2018)}]{yatskar2018qualitative}
Mark Yatskar. 2018.
\newblock A qualitative comparison of coqa, squad 2.0 and quac.
\newblock \emph{arXiv preprint arXiv:1809.10735}.

\bibitem[{Yu et~al.(2018)Yu, Dohan, Luong, Zhao, Chen, Norouzi, and
  Le}]{yu2018qanet}
Adams~Wei Yu, David Dohan, Minh-Thang Luong, Rui Zhao, Kai Chen, Mohammad
  Norouzi, and Quoc~V Le. 2018.
\newblock Qanet: Combining local convolution with global self-attention for
  reading comprehension.
\newblock \emph{arXiv preprint arXiv:1804.09541}.

\bibitem[{Zhu et~al.(2018)Zhu, Zeng, and Huang}]{zhu2018sdnet}
Chenguang Zhu, Michael Zeng, and Xuedong Huang. 2018.
\newblock Sdnet: Contextualized attention-based deep network for conversational
  question answering.
\newblock \emph{arXiv preprint arXiv:1812.03593}.

\end{thebibliography}
\bibliographystyle{acl_natbib}

\clearpage
\appendix

\section{Reducing Sequential Intruction Understanding to Conversational Machine Comprehension} \label{sec:scone_reduce}

\begin{figure}[h!]
  \centering
  \includegraphics[width=\linewidth]{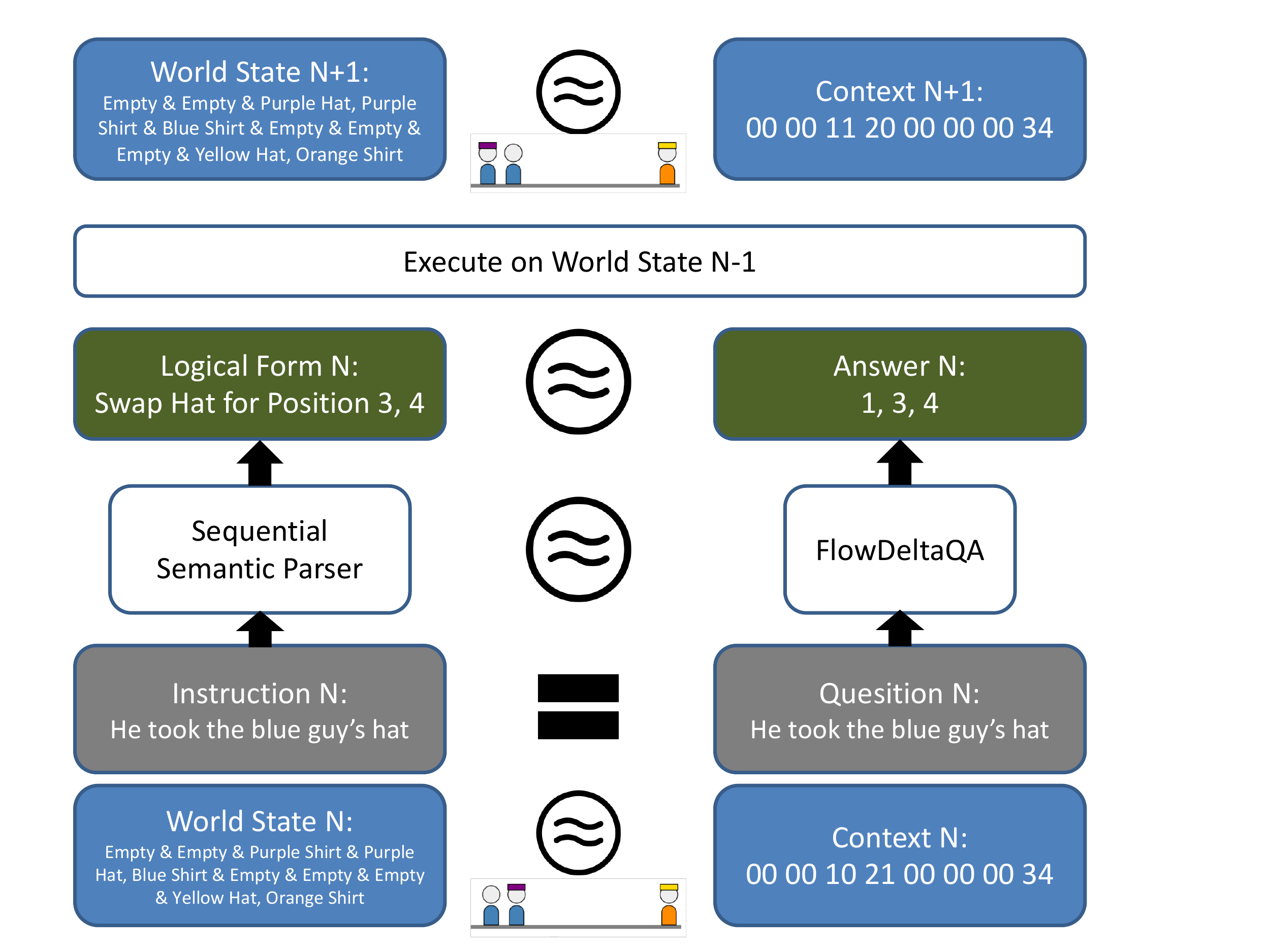}
  \caption{Example of the SCONE dataset and its reduction }
  \label{fig:scone}
  \vspace{-2mm}
\end{figure}

In SCONE dataset, given the initial world state $\boldsymbol{W}_0$ and a sequence of natural language insturctions $\{\boldsymbol{I_1, \ldots \boldsymbol{I_K}}\}$, the model need to perform the correct sequence of actions on $\boldsymbol{W}_0$ and obtain the correct world states $\{\boldsymbol{W_1}, \ldots, \boldsymbol{W_K}\}$ after each instruction. 
An example from \cite{long2016simpler} is shown in the left-hand side of Figure \ref{fig:scone}.

Following FlowQA \cite{huang2018flowqa}, for each position in the world state, we encode it as two integers denoting the shirt and hat color in \textsc{Scene}, image ID and present or not in \textsc{Tangrams}, and color of the liquid and number of units in \textsc{Alchemy}. 
Next, the change of world states (i.e., the logical form) is encoded as three or four integers. The first integers is the type of action performed. The second and third integers represent the position of the context (i.e., the encoded world state). Finally, the fourth integer represents the additional property for the action such as the number of units moved. 

An example of encoded world states and logical form is shown in the right-hand side of Figure \ref{fig:scone}. In this example, action $(1, 3, 4)$ means "swap the hat for position 3, 4" and there is no additional property for the action. 

\section{Experimental Details}
We reproduce and report the experiment results of FlowQA using the released code except SCONE part since the official released code does not contain it. 
Authors claim there is further performance improvement on the released version of FlowQA

All hyperparameters are kept the same as recommended one in FlowQA and BERT for CoQA and QuAC datasets.
For SCONE, due to the relatively small size of dataset, to prevent overfitting we further tune the hidden size of \textsc{FlowDeltaQA} in three different domains. 
The tuned hidden sizes are $50, 60, 70$ for \textsc{Scene}, \textsc{Alchemy} and \textsc{Tangrams} respectively.

\section{Flow Information Gain Variants} \label{sec: flow_variant}

We test three different variants of \textsc{FlowDelta} on modeling the information flow in the dialog and show results in table \ref{tab: coqa_flow_variant}. The three variants are:
\begin{enumerate}
    \item SkipDelta: $h_{t-1} - h_{t-3}$
    \item DoubleDelta: $[h_{t-1} - h_{t-2}; h_{t-2} - h_{t-3}]$
    \item Hadamard Product: $h_{t-1} * h_{t-2}$
\end{enumerate}
The reason to use SkipDelta and DoubleDelta is because we want to see if there is any benefit to incorporate longer (or more) dialog history. 
Experiment results show while using longer dialog history (i.e., SkipDelta) helps, adding too many dialog history (i.e., DoubleDelta) does not give any improvement.

The intuition behind Hadamard product is to model the similarity of consecutive hidden states. 
If there are any topic shift in last turns of dialog, we expect Hadamard product can give us useful signal to detect it.
Results show although the proposed \textsc{FlowDelta} is the best, Hadamard product outperforms SkipDelta and DoubleDelta and proves its effectiveness.

\begin{table}[h!]
    \begin{center}
    \begin{tabular}{|l|c|}
    \hline
        {\bf Model} & {\bf F1} \\
    \hline\hline
        FlowQA & 76.7 \\
        FlowDeltaQA (SkipDelta) & 76.9 \\
        FlowDeltaQA (DoubleDelta)& 76.7 \\
        FlowDeltaQA (Hadamard Product) & 77.2 \\
        FlowDeltaQA  & \bf 77.6 \\
    \hline
    \end{tabular}
    \end{center}
    \caption{\label{tab: coqa_flow_variant} CoQA results of different variants of \textsc{flow} interaction. All models are provided with previous 1 gold answer.}
\end{table}

\section{Qualitative Analysis}
\label{sec:qual}
\begin{table*}[h!]
    \begin{center}
    \begin{tabular}{|l|p{2.8cm}p{3.5cm}p{3.5cm}|}
    \hline
    \bf Questions & \bf FlowQA & \bf FlowDeltaQA & \bf Gold Answer \\
    \hline
    Whose house was searched? &  \multicolumn{3}{c|}{Gary Giordano} \\
   In what city? & \multicolumn{3}{c|}{Gaithersburg} \\
   County? &  \multicolumn{3}{c|}{Montgomery County} \\
   State? &  \multicolumn{3}{c|}{Maryland} \\
   Where is he now? &  \multicolumn{3}{c|}{Aruban jail}\\
   Why? & lack of evidence  & 6 recent disappearance of an American woman & suspect in the recent disappearance of an American woman\\
    \hline
    \end{tabular}
    \end{center}
    \caption{\label{tab: qualitative_analysis} Qualitative analysis of FlowDeltaQA. }
\end{table*}

Here we present an example from CoQA dataset which consists of a passage that the dialog talks about, and a sequence of questions and answers.
Table \ref{tab: qualitative_analysis} shows the questions, answers and model predicitons. 
We note the gold answer in CoQA is abstractive and may not be a span in the passage.
Only a subset of the dialog is showed to demonstrate the different behaviors of FlowQA and FlowDeltaQA.

\textbf{Context}: 
(CNN) -- FBI agents on Friday night searched the Maryland home of the suspect in the recent disappearance of an American woman in Aruba, an agent said. 
The search is occurring in the Gaithersburg residence of Gary Giordano, who is currently being held in an Aruban jail, FBI Special Agent Rich Wolf told CNN. 
Agents, wearing vests that said FBI and carrying empty cardboard and plastic boxes, arrived about 8:40 p.m. Friday. About 15 unmarked cars could be seen on the street, as well as a Montgomery County police vehicle. 
Supervisory Special Agent Philip Celestini, who was at the residence, declined to comment further on the search, citing the active investigation. 
Aruban Solicitor General Taco Stein said earlier Friday that the suspect will appear in court Monday, where an investigating magistrate could order him held for at least eight more days, order him to remain on the island or release him outright due to a lack of evidence. 
Giordano was arrested by Aruban police on August 5, three days after Robyn Gardner was last seen near Baby Beach on the western tip of the Caribbean island. 
Giordano told authorities that he had been snorkeling with Gardner when he signaled to her to swim back, according to a statement. When he reached the beach, Gardner was nowhere to be found, Giordano allegedly said. 
The area that Giordano led authorities to is a rocky, unsightly location that locals say is not a popular snorkeling spot. 
Although prosecutors have continued to identify the 50-year-old American man by his initials, GVG, they also released a photo of a man who appears to be Giordano. His attorney, Michael Lopez, also has said that his client is being held as a suspect in Gardner's death. Lopez has not returned telephone calls seeking comment. 

\paragraph{Analysis} 
In this example, to answer the last question "Why?", model need to understand the previous conversation correctly to know the actual question is "Why is Gary Giordano in the Aruban Jail now?".
This example is particularly hard since in order to know "he" in "Where is he now?" refers to "Gary Giordano", model need to use the information from the very first question "Whose house was searched", which requires the ability to utilize full dialog history.
While \textsc{FlowQA} fails to hook this question to the correct conversation context and respond reasonable but incorrect answer, our \textsc{FlowDeltaQA} successfully grasps long dialog flow and answers the correct span. 

\end{document}